\documentclass[11pt]{article}

\usepackage[a4paper,margin=1in]{geometry}
\usepackage{times}
\usepackage{natbib}

\usepackage{amsmath,amsfonts,bm}

\def\eqref#1{equation~\ref{#1}}
\def\1{\bm{1}}

\DeclareMathAlphabet{\mathsfit}{\encodingdefault}{\sfdefault}{m}{sl}
\SetMathAlphabet{\mathsfit}{bold}{\encodingdefault}{\sfdefault}{bx}{n}

\usepackage{hyperref}
\usepackage{url}
\usepackage{booktabs}
\usepackage{amsmath}
\usepackage{amssymb}
\usepackage{amsfonts}
\usepackage{multirow}
\usepackage{algorithm}
\usepackage{algorithmic}
\usepackage{array}
\usepackage{tabularx}
\usepackage{makecell}
\usepackage{graphicx}

\title{Rethinking EEG Spatial Super-Resolution as Dynamic State Decomposition and Filtering}

\author{
Hongjun Liu\textsuperscript{1},
Leyu Zhou\textsuperscript{1},
Chang Ge\textsuperscript{1},
Shitong Duan\textsuperscript{1},
Chao Yao\textsuperscript{1}
\\[0.5em]
\small \textsuperscript{1}University of Science and Technology Beijing
}

\date{}

\begin{document}

\maketitle

\begin{abstract}
EEG spatial super-resolution aims to recover high-density neural signals from sparse electrode observations. 
Existing methods have achieved substantial progress through spatial modeling, generative reconstruction, and temporal representation learning.
However, the current EEG state contains two different sources of information: predictable evolution inherited from preceding states and newly emerging changes revealed by the current observation. Most approaches typically do not model these two components explicitly.
In this work, we formulate streaming EEG spatial super-resolution as a sequential state filtering problem and propose Dynamic State Filtering (DSF). 
Rather than relying on a single temporal reconstruction pathway, DSF combines a history-aware dynamical branch with an innovation-filtering branch that explicitly decomposes waveform-state estimation into prediction and observation-driven correction. To infer how spatially incomplete innovations should affect missing locations, DSF routes the observed innovations across electrodes through a mask-constrained, topology-aware transition operator and incorporates them through uncertainty-aware state correction.
The two complementary estimates are finally fused to reconstruct the dense EEG signal.
Experiments on SEED, SEED-IV and BCI2000 demonstrate that DSF improves reconstruction fidelity and downstream utility while maintaining robustness under dynamically changing and unreliable sparse observations.
\end{abstract}

\section{Introduction}

High-density electroencephalography (EEG) provides fine-grained spatial measurements of neural activity, but dense electrode recordings remain costly and inconvenient to acquire because of setup complexity, wearing comfort, and sensing reliability \citep{tang2023deep,li2025estformer}. EEG spatial super-resolution (EEGSR) therefore aims to recover high-density signals from sparse electrode observations. Beyond spatial missingness, EEGSR also involves a temporal estimation problem: consecutive EEG segments are not independent, and part of the current neural activity is predictable from preceding states, while newly observed measurements provide additional evidence that cannot be inferred from history alone \citep{mckenna1994brain}.

\begin{figure}[t]
\centering
  \includegraphics[width=0.9\columnwidth]{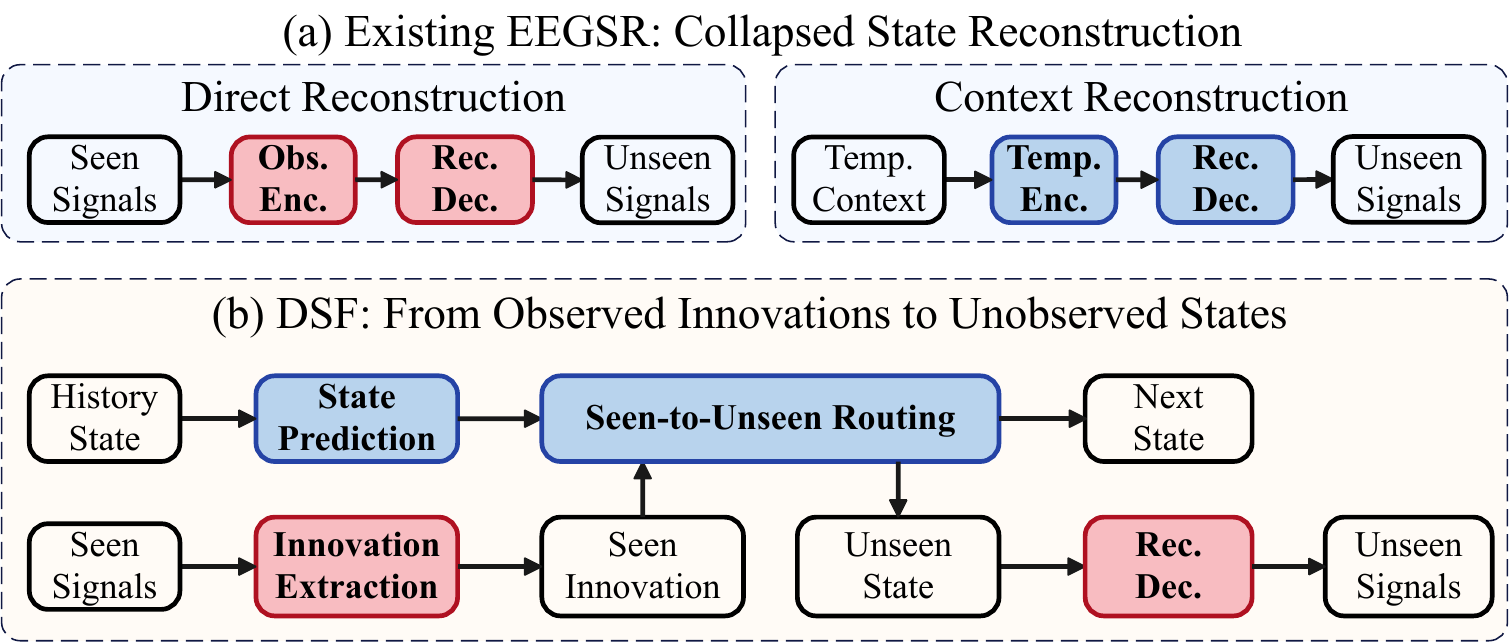}
  \caption{Motivation of DSF. Conventional segment-wise EEG spatial super-resolution estimates each target dense EEG segment primarily from the corresponding sparse observation or temporally aggregated features. DSF instead separates two complementary sources of information: historical dynamics provide a prediction of the current EEG state, while the current sparse observation contributes innovations that correct this prediction at both observed and unobserved electrodes. 
}
  \label{fig:motivation}
\end{figure}

Existing EEGSR approaches exploit interpolation, spatial dependencies, generative priors, and temporal modeling to infer dense EEG from sparse measurements \citep{elouahidi2023spatial,maser2024enhancing,liu2026srgdiff,liu2026cafe}. However, temporal context is typically incorporated as an additional representation for direct reconstruction. The reconstruction process does not explicitly distinguish a state predicted from preceding EEG from an observation-induced correction that is available only at currently visible electrodes. This distinction becomes particularly important when the current measurement is sparse, corrupted, or temporarily unavailable: historical information can provide a useful prior, but blindly incorporating unreliable current measurements may also distort the estimated state.

This motivates us to formulate streaming EEG spatial super-resolution as sequential state filtering. Rather than asking a model to regenerate the complete dense signal at every step solely from the current sparse input, we maintain an evolving state across consecutive EEG windows. Each update therefore contains two distinct sources of information: a predictive prior propagated from the preceding state and an innovation revealed by the current observation. However, because the current observation is spatially sparse, this innovation can be directly computed only at visible electrodes; no observation-derived innovation is available at missing locations. The key problem is therefore not merely how to exploit temporal dependencies, but how innovations observed at a subset of electrodes should correct the predicted states of both observed and unobserved electrodes.

To realize this formulation, we propose \textbf{Dynamic State Filtering (DSF)}, which combines two complementary causal pathways. A history-aware dynamical branch propagates latent states across consecutive EEG windows and updates them with the current sparse observation, while an innovation-filtering branch explicitly decomposes waveform-state estimation into prediction and observation-driven correction. Rather than propagating raw observations or generic hidden features, the filtering branch treats the observed deviations as innovation sources and routes them toward both observed and unobserved states through a mask-constrained, topology-aware transition operator. The two branches are finally integrated to produce the dense EEG reconstruction.

Our contributions are summarized as follows:
\begin{itemize}
    \item We formulate streaming EEG spatial super-resolution as causal state filtering across consecutive EEG windows, explicitly separating a history-derived predictive state from corrections introduced by the current sparse observation.
    \item We develop a dual-branch DSF framework that combines causal dynamical modeling with mask-constrained, topology-aware innovation routing, allowing current innovations from visible electrodes to correct unobserved states.
    \item We evaluate DSF on three EEG datasets across multiple sparsity levels and observation perturbations, together with downstream task evaluation and component analyses, demonstrating consistent advantages over static reconstruction and temporal modeling baselines.
\end{itemize}

\section{Related Work}

\begin{figure*}[t]
  \centering
  \includegraphics[width=1.0\columnwidth]{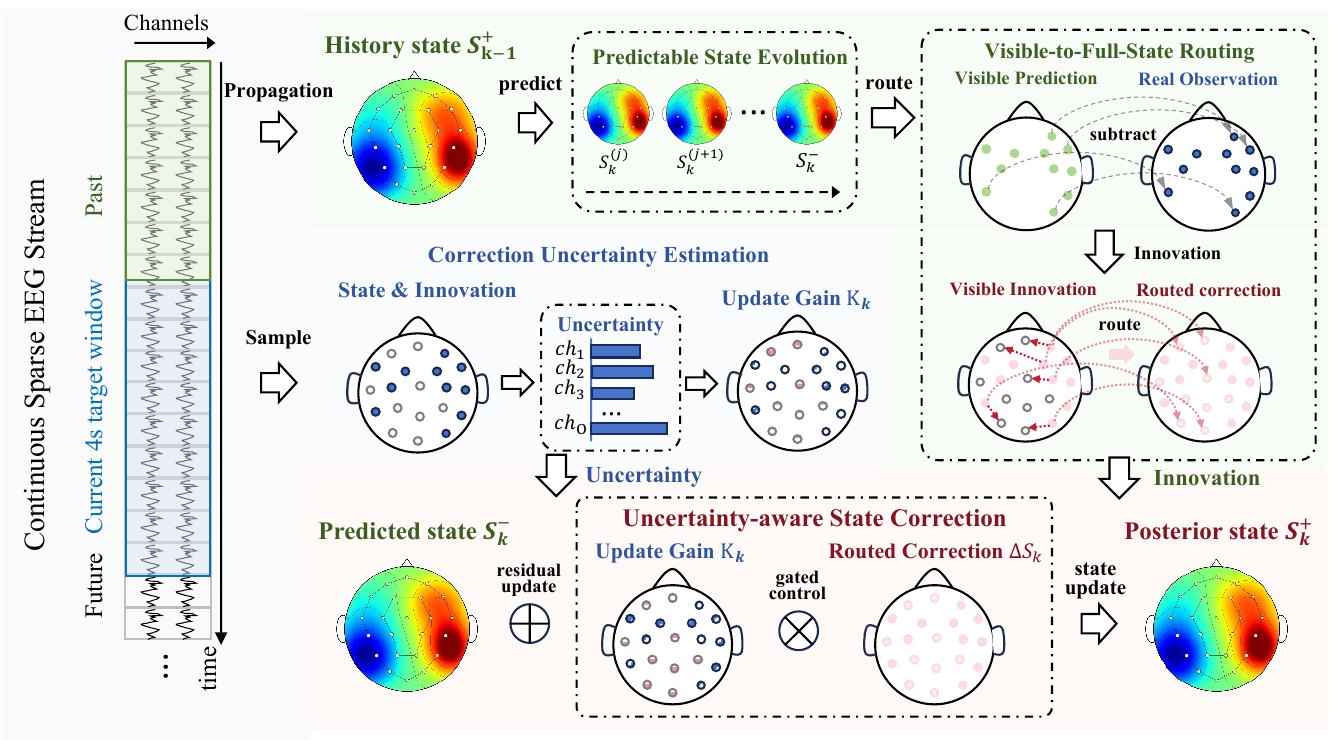}
  % \captionsetup{width=1.9\columnwidth}
\caption{Overview of the innovation-filtering branch of DSF. The previous waveform state is propagated to obtain a predictive state, while innovations extracted from the current sparse observation are routed from observed to unobserved electrodes through a state-conditioned, topology-aware transition operator. The routed correction is then incorporated through uncertainty-aware state updating to obtain the posterior state.
}
  \label{fig:framework}
\end{figure*}
EEG spatial super-resolution aims to recover high-density (HD) EEG from sparse low-density (LD) observations. Early studies relied primarily on interpolation or adversarial generation, such as EEGSR-GAN \cite{corley2018deep}, while recent approaches incorporate spatial connectivity, Transformer architectures, diffusion models, and channel-wise dependency modeling \cite{tang2023deep,wang2025stad,li2025estformer,liu2026srgdiff}. Temporal dynamics have also been introduced into EEGSR; for example, MASER employs state-space modeling to capture temporal dependencies in EEG reconstruction \cite{maser2024enhancing}. Our focus is therefore not temporal modeling itself, but how a causal state propagated across consecutive EEG windows should be corrected when current observations are spatially incomplete. DSF represents this correction as innovation observed only at visible electrodes and routes these innovations across the electrode topology before updating the predicted state.

Sequential state estimation provides the conceptual basis for this prediction--innovation formulation. Classical Kalman filtering performs sequential estimation through alternating prediction and innovation-based measurement updates \cite{kalman1960new}. Neural state-space approaches extend this principle to nonlinear and irregular dynamics, including Neural ODEs, Latent ODEs, and Neural CDEs \cite{chen2018neural,rubanova2019latent,kidger2020neuralcde}, while S4, S5, and Mamba provide powerful learned sequence models \cite{gu2022s4,smithsimplified,dao2024mamba}. Graph-based imputation methods such as GRIN further demonstrate that information can be propagated across correlated variables under incomplete observations \citep{cini2022grin}. DSF adopts the general prediction--correction principle. We extend it to partially observed spatial state estimation, where the current state is observed only at a subset of locations and the resulting innovations must be propagated from observed to unobserved locations before updating the full state.

\section{Method}

\subsection{Problem Formulation and Overview}
\label{sec:method_overview}

We consider EEG spatial super-resolution over a temporally ordered sequence of EEG segments. Let $X_k\in\mathbb{R}^{C\times T}$ denote the full-density EEG segment at step $k$, where $C$ and $T$ are the numbers of electrodes and temporal samples, respectively. Only a subset of electrodes is available at each step. We denote the corresponding visibility mask by $M_k\in\{0,1\}^{C}$ and construct the zero-filled sparse observation as $O_k=M_k\odot X_k$, where the mask is broadcast along the temporal dimension. Given observations up to the current step, $\{(O_i,M_i)\}_{i=0}^{k}$, the goal is to reconstruct $X_k$ without accessing future EEG segments.

DSF approaches this problem through two complementary causal pathways.
The dynamical branch performs history-aware temporal reconstruction by
propagating a compact latent state across consecutive EEG windows and updating it with the current sparse observation, producing a dense estimate $\hat X_k^{\mathrm{dyn}}$.
In parallel, the innovation-filtering branch explicitly decomposes waveform-state estimation into prediction and observation-driven correction. It first propagates the previous waveform state to obtain a predictive prior, computes the innovation introduced by the current sparse observation only at visible electrodes, and routes this innovation to the full scalp before state correction, yielding $\hat X_k^{\mathrm{filt}}$. DSF finally combines their predictions through a bounded channel-wise fusion. Figure~\ref{fig:framework} illustrates the innovation-filtering branch in detail.

\subsection{History-driven Dynamical Prediction}
\label{sec:dynamical_prediction}

Existing temporal EEGSR approaches model temporal dependencies by aggregating or propagating representations across EEG segments.
In contrast, DSF first propagates the historical state to obtain a predictive representation, which is subsequently updated by the current sparse observation.
Specifically, the dynamical branch jointly encodes the sparse waveform and its visibility mask into a latent observation $Z_k\in\mathbb{R}^{D_h\times T_h}$, avoiding ambiguity between unobserved electrodes and genuine low-amplitude measurements.

The first latent observation initializes the recurrent state. For each subsequent EEG window, the previous updated state is first propagated through a learned continuous dynamics function and then corrected by the encoded current observation:
\begin{equation}
H_k^-=\Phi_\theta(H_{k-1}^+),\qquad
H_k^+(:,\tau)
=
\operatorname{GRU}
\bigl(Z_k(:,\tau),H_k^-(:,\tau)\bigr),
\qquad
\hat X_k^{\mathrm{dyn}}
=
\mathcal D_{\mathrm{dyn}}(H_k^+).
\label{eq:dynamical_prediction}
\end{equation}
Here, $H_k^-,H_k^+\in\mathbb{R}^{D_h\times T_h}$ denote the propagated and updated latent states, respectively, and the GRU parameters are shared across latent temporal positions. $\Phi_\theta$ models continuous latent-state evolution, while $\mathcal D_{\mathrm{dyn}}$ maps the updated state back to waveform space. Detailed encoder, dynamics, decoder, and numerical integration
configurations are provided in Appendix~\ref{app:model_details}.

This branch therefore yields a dense estimate conditioned primarily on the history of the EEG stream. However, temporal prediction alone cannot account for neural changes that first appear in the current measurement. We next isolate these observation-induced changes and determine how they should affect electrodes that are currently unobserved.

\subsection{Topology-aware Innovation Routing}
\label{sec:innovation_routing}

Given the history-driven prediction, we next extract the information newly introduced by the current observation.
The filtering branch maintains a waveform-domain state $S_k$ and propagates the previous updated state through a shared temporal dynamics function. Specifically, starting from $S_k^{(0)}=S_{k-1}^+$, we perform $N_{\mathrm{E}}$ Euler updates:
\begin{equation}
S_k^{(s+1)}
=
S_k^{(s)}
+
\frac{d_t}{N_{\mathrm{E}}}
f_{\omega}\!\left(S_k^{(s)}\right),
\qquad
s=0,\ldots,N_{\mathrm{E}}-1,
\end{equation}
and set $S_k^-=S_k^{(N_{\mathrm{E}})}$.
Here, $f_{\omega}$ is an electrode-shared temporal velocity field, and we use $N_{\mathrm{E}}=4$ and a normalized propagation interval $d_t=1.0$.
Since $S_k^-$ and $O_k$ lie in the same waveform space, their discrepancy
$E_k=O_k-M_k\odot S_k^-$ directly represents the innovation available at currently visible electrodes.

The central difficulty is that $E_k$ is observed only at a subset of locations, whereas the correction must update the entire scalp. We therefore condition spatial propagation jointly on the predictive waveform state, the observed innovation, and electrode topology. Specifically, $S_k^-$ and $E_k$ are encoded independently for each electrode by temporal encoders shared across electrodes, producing $F_k^{p},F_k^{e}\in\mathbb{R}^{C\times d\times T}$. Temporal aggregation then produces electrode-wise tokens, from which the predictive query and innovation key are obtained as
\begin{equation}
q_{k,i}
=
W_q\operatorname{Avg}_{t}(F_{k,i}^{p}),
\qquad
\kappa_{k,j}
=
W_k\operatorname{Avg}_{t}(F_{k,j}^{e}),
\end{equation}
while the temporally resolved innovation representation
$V_{k,j}
=
\operatorname{LN}(F_{k,j}^{e})
\in\mathbb{R}^{d\times T}$
serves as the routed value.
Let $B_{ij}=-d_{ij}^{2}/(2\sigma_p^{2})$ denote the topology bias derived from electrode distance. Innovation routing and aggregation are then performed as
\begin{equation}
A_k(i,j)
=
\frac{
M_k(j)\exp\!\left(
q_{k,i}^{\top}\kappa_{k,j}/\sqrt{d}
+
B_{ij}
\right)
}{
\sum_l
M_k(l)\exp\!\left(
q_{k,i}^{\top}\kappa_{k,l}/\sqrt{d}
+
B_{il}
\right)
},
\qquad
\Delta S_k(i,:)
=
\phi_v\!\left(
\sum_j A_k(i,j)V_{k,j}
\right).
\label{eq:innovation_routing}
\end{equation}
Here, $A_k(i,j)$ measures how strongly the innovation observed at electrode $j$ contributes to correcting electrode $i$, while the visibility mask prevents unobserved electrodes from acting as innovation sources.
The aggregated innovation is then projected back to waveform space to form the candidate correction $\Delta S_k$.

The routing operator thus answers \emph{where} the currently observed innovation should propagate. It does not yet determine \emph{how strongly} the resulting correction should modify the predictive state, particularly when the current observation is noisy or inconsistent with the propagated state. We address this next through an uncertainty-aware state update.

\subsection{Uncertainty-aware State Correction}
\label{sec:uncertainty_correction}

After routing, $\Delta S_k$ provides a candidate correction for every electrode. Applying it directly would implicitly assign the same confidence to all observation-driven updates. Instead, the filtering branch maintains a channel-wise uncertainty state $P_k\in\mathbb{R}^{C}$ and uses it to regulate how strongly the routed correction is written into the waveform state.
DSF propagates the previous uncertainty to obtain $P_k^-$ and estimates a correction uncertainty $R_k$ from the predictive state, the current innovation, and the routed correction $\Delta S_k$. Specifically, the propagated uncertainty is
$P_k^- = P_{k-1}^+ + Q$, where $Q\in\mathbb{R}^{C}$ is a positive
learnable channel-wise process term. The correction uncertainty is estimated
as
\begin{equation}
R_k = R_{\mathrm{floor}} +
\operatorname{softplus}\!\left(
h_R[
\bar F_k^p;\bar F_k^e;
\operatorname{Avg}_t(\Delta S_k);
\operatorname{Avg}_t|\Delta S_k|
]
\right),
\end{equation}
where
$\bar F_k^p=\operatorname{Avg}_t(F_k^p)$ and
$\bar F_k^e=\operatorname{Avg}_t(F_k^e)$
are temporally averaged predictive-state and innovation features,
$h_R$ is a two-layer MLP shared across electrodes, and
$R_{\mathrm{floor}}\in\mathbb{R}^{C}$ is a positive learnable floor.
The relative uncertainties of the propagated state and the new correction determine a bounded update gain:
\begin{equation}
K_k
=
\frac{P_k^-}{P_k^-+R_k+\epsilon},
\qquad
S_k^+
=
S_k^-+K_k\odot\Delta S_k,
\qquad
P_k^+
=
(1-K_k)\odot P_k^- .
\label{eq:uncertainty_update}
\end{equation}
Consequently, corrections with larger estimated uncertainty have less
influence on the propagated state, while more certain observations induce stronger updates.
We set $\epsilon=10^{-6}$ for numerical stability and treat $P_k$ and $R_k$ as learned uncertainty surrogates rather than calibrated Bayesian covariance estimates.
The updated waveform state is decoded into the filtering estimate $\hat X_k^{\mathrm{filt}}=\mathcal D_{\mathrm{filt}}(S_k^+)$. At this point, DSF has obtained two complementary reconstructions: one dominated by history-driven latent dynamics and one that explicitly incorporates current observation innovations.

\subsection{Bounded Fusion and Learning}
\label{sec:fusion_learning}

Because the two branches capture complementary information, we do not force either estimate to explain the complete reconstruction alone. Instead, DSF learns one bounded fusion coefficient for each electrode. The final prediction is
\begin{equation}
\alpha_c=\alpha_{\max}\sigma(a_c),\qquad
\hat X_k(c,:)
=
(1-\alpha_c)\hat X_k^{\mathrm{dyn}}(c,:)
+
\alpha_c\hat X_k^{\mathrm{filt}}(c,:),
\label{eq:bounded_fusion}
\end{equation}
where $\alpha_c$ is a bounded electrode-specific coefficient shared across waveform time.
This design preserves the history-driven estimate as the primary prediction while allowing the innovation-filtering branch to provide adaptive channel-wise corrections.

Training follows the same decomposition. The two branches are first optimized independently using waveform reconstruction, visible-channel consistency, and STFT-based spectral preservation objectives, with an additional uncertainty regularization applied to the filtering branch. Their parameters are then frozen, and only the channel-wise fusion coefficients are optimized. Detailed loss definitions and optimization settings are provided in Appendix~\ref{app:implementation}.

\section{Experiments}

\begin{table*}[t]
\centering
\footnotesize
\setlength{\tabcolsep}{3.8pt}

\begin{tabular}{l|ccc|ccc|ccc}
\toprule
\multirow{2}{*}{Method}
& \multicolumn{3}{c|}{SEED}
& \multicolumn{3}{c|}{SEED-IV}
& \multicolumn{3}{c}{BCI2000}\\
&
NMSE$\downarrow$
& PCC$\uparrow$
& SNR$\uparrow$
& NMSE$\downarrow$
& PCC$\uparrow$
& SNR$\uparrow$
& NMSE$\downarrow$
& PCC$\uparrow$
& SNR$\uparrow$\\
\midrule

%%%%%%%%%%%%%%%%%%%%%%%%%%%%%%%%%%%%%%%%%%%%%%%%%%%%%%%%%%%%
\multicolumn{10}{c}{\textbf{Scale factor = $2\times$}}\\
\midrule

Spherical Spline
& 0.2912 & 0.8175 & 5.9821
& 0.3026 & 0.8108 & 5.5473
& 0.1269 & 0.9251 & 10.0374\\

ESTFormer
& 0.2531 & 0.8324 & 7.1178
& 0.2323 & 0.8275 & 7.3576
& 0.0876 & 0.9432 & 11.9237\\

STAD
& 0.2344 & 0.8582 & 7.2856
& 0.2291 & 0.8587 & 7.3870
& 0.0884 & 0.9401 & 11.8492\\

SRGDiff
& 0.2749 & 0.8588 & 7.0047
& 0.2229 & 0.8564 & 7.4213
& 0.0898 & 0.9412 & 11.8250\\

\midrule

MASER
& 0.2387 & 0.8551 & 7.2036
& 0.2768 & 0.8281 & 5.9462
& 0.0891 & 0.9410 & 11.8015\\

Mamba aggregation
& 0.2441 & 0.8576 & 6.5998
& 0.2699 & 0.8391 & 6.0909
& 0.1076 & 0.9477 & 11.8853\\

Kalman Filter
& 0.2568 & 0.8421 & 6.2144
& 0.2815 & 0.8246 & 5.8231
& 0.1178 & 0.9342 & 10.3215\\

Latent ODE
& 0.2244 & 0.8603 & 6.9757
& 0.2522 & 0.8516 & 6.4223
& 0.1029 & 0.9466 & 11.6573\\

\midrule

DSF
& \textbf{0.2009}
& \textbf{0.8858}
& \textbf{7.5341}
& \textbf{0.2120}
& \textbf{0.8766}
& \textbf{7.4504}
& \textbf{0.0636}
& \textbf{0.9683}
& \textbf{13.8554}\\

%%%%%%%%%%%%%%%%%%%%%%%%%%%%%%%%%%%%%%%%%%%%%%%%%%%%%%%%%%%%
\midrule
\multicolumn{10}{c}{\textbf{Scale factor = $4\times$}}\\
\midrule

Spherical Spline
& 0.5784 & 0.7116 & 3.6815
& 0.5389 & 0.6483 & 2.8764
& 0.1907 & 0.8598 & 8.1032\\

ESTFormer
& 0.5206 & 0.7455 & 4.0474
& 0.4113 & 0.7383 & 4.2234
& 0.1653 & 0.8911 & 9.3554\\

STAD
& 0.5511 & 0.7439 & 4.0344
& 0.4151 & 0.7379 & 4.3070
& 0.1670 & 0.8920 & 9.2867\\

SRGDiff
& 0.4535 & 0.7528 & 4.2195
& 0.4174 & 0.7459 & 4.2758
& 0.1641 & 0.8976 & 9.3285\\

\midrule

MASER
& 0.4342 & 0.7354 & 3.9486
& 0.4193 & 0.7362 & 4.2541
& 0.1708 & 0.8742 & 8.5037\\

Mamba aggregation
& 0.4397 & 0.7508 & 4.1536
& 0.4886 & 0.6832 & 3.3386
& 0.1669 & 0.8930 & 8.7002\\

Kalman Filter
& 0.4316 & 0.7325 & 3.9214
& 0.5127 & 0.6648 & 3.1127
& 0.1715 & 0.8713 & 8.4126\\

Latent ODE
& 0.4418 & 0.7289 & 3.9044
& 0.4570 & 0.7116 & 3.6276
& 0.1792 & 0.8866 & 8.5519\\

\midrule

DSF
& \textbf{0.3875}
& \textbf{0.7812}
& \textbf{4.4382}
& \textbf{0.3925}
& \textbf{0.7593}
& \textbf{4.4527}
& \textbf{0.1334}
& \textbf{0.9251}
& \textbf{10.5268}\\

%%%%%%%%%%%%%%%%%%%%%%%%%%%%%%%%%%%%%%%%%%%%%%%%%%%%%%%%%%%%
\midrule
\multicolumn{10}{c}{\textbf{Scale factor = $8\times$}}\\
\midrule

Spherical Spline
& 0.8126 & 0.4998 & 1.4763
& 0.6954 & 0.5312 & 1.8217
& 0.2641 & 0.8195 & 6.4038\\

ESTFormer
& 0.7815 & 0.5318 & 1.8751
& 0.5756 & 0.6348 & 2.5792
& 0.2375 & 0.8446 & 6.7209\\

STAD
& 0.7260 & 0.5489 & 1.8050
& 0.5747 & 0.6259 & 2.5906
& 0.2416 & 0.8327 & 6.6454\\

SRGDiff
& 0.7092 & 0.5409 & 1.8957
& 0.5736 & 0.6265 & 2.4943
& 0.2380 & 0.8442 & 6.8003\\

\midrule

MASER
& 0.7318 & 0.5467 & 1.8114
& 0.6581 & 0.5527 & 2.0819
& 0.2428 & 0.8341 & 6.7016\\

Mamba aggregation
& 0.7163 & 0.5367 & 1.8561
& 0.6317 & 0.5676 & 2.1526
& 0.2497 & 0.8529 & 7.8285\\

Kalman Filter
& 0.7512 & 0.5184 & 1.6723
& 0.6645 & 0.5489 & 2.0134
& 0.2518 & 0.8446 & 7.4217\\

Latent ODE
& 0.7187 & 0.5377 & 1.7848
& 0.5963 & 0.6045 & 2.4095
& 0.2443 & 0.8435 & 7.5201\\

\midrule

DSF
& \textbf{0.6672}
& \textbf{0.5854}
& \textbf{2.1514}
& \textbf{0.5346}
& \textbf{0.6610}
& \textbf{2.9690}
& \textbf{0.2075}
& \textbf{0.8846}
& \textbf{8.1209}\\

\bottomrule
\end{tabular}

\caption{
Main reconstruction results under the clean-layout protocol across different reconstruction scales.
NMSE, PCC, and SNR are computed exclusively on electrodes absent from the sparse input; visible electrodes are not replaced by ground-truth signals during primary metric computation.
}
\label{tab:main_results}
\end{table*}

\begin{table*}[h]
\centering
\footnotesize
\begin{tabular}{ll|cc|cc|cc}
\toprule
\multirow{2}{*}{Category}
& \multirow{2}{*}{Method}
& \multicolumn{2}{c|}{Noisy Observation}
& \multicolumn{2}{c|}{Channel Dropout}
& \multicolumn{2}{c}{Noise + Dropout}\\
&
& NMSE$\downarrow$ & PCC$\uparrow$
& NMSE$\downarrow$ & PCC$\uparrow$
& NMSE$\downarrow$ & PCC$\uparrow$\\
\midrule

\multirow{3}{*}{Static EEG SR}
& ESTFormer
& 0.5469 & 0.7503
& 0.5959 & 0.6976
& 0.6148 & 0.6838\\

& STAD
& 0.5945 & 0.7411
& 0.6385 & 0.6850
& 0.6690 & 0.6653\\

& SRGDiff
& 0.4906 & 0.7524
& 0.6918 & 0.6305
& 0.7237 & 0.6134\\

\midrule

\multirow{3}{*}{Temporal Modeling}
& Mamba Aggregation
& 0.4369 & 0.7504
& 0.5251 & 0.6817
& 0.5516 & 0.6812\\

& Kalman Filter
& 0.4685 & 0.7246
& 0.5568 & 0.6521
& 0.5862 & 0.6435\\

& Latent ODE
& 0.4493 & 0.7238
& 0.5379 & 0.6759
& 0.5579 & 0.6693\\

\midrule

\multirow{2}{*}{Ours}

& DSF w/o uncertainty
& 0.4281
& 0.7455
& 0.5189
& 0.6834
& 0.5513
& 0.6785\\

& DSF
& \textbf{0.4014} & \textbf{0.7727}
& \textbf{0.4933} & \textbf{0.7078}
& \textbf{0.5048} & \textbf{0.6998}\\

\bottomrule
\end{tabular}
\caption{ Robustness under observation uncertainty under the $4\times$ setting. We evaluate noisy visible channels, additional channel dropout, and their combination under the same setting as the corresponding clean results in Table~\ref{tab:main_results}. }
\label{tab:robustness_uncertainty}
\end{table*}

\begin{table}[t]
\centering
\footnotesize
% \resizebox{\columnwidth}{!}{
\begin{tabular}{l|cc|cc|cc}
\toprule
Method
& \multicolumn{2}{c|}{SEED}
& \multicolumn{2}{c|}{SEED-IV}
& \multicolumn{2}{c}{BCI2000}\\
& Acc & F1
& Acc & F1
& Acc & F1\\
\midrule

Sparse LD
&33.33&30.22
&25.00&13.11
&27.78&22.49\\

GT-HD
&37.87&34.13
&30.42&17.96
&45.71&44.98\\

\midrule

ESTFormer
&34.82&32.34
&28.14&15.07
&42.84&40.89\\

STAD
&35.40&31.72
&28.01&15.68
&42.87&40.99\\

SRGDiff
&35.96&32.50
&28.21&15.69
&42.68&40.72\\

Mamba aggregation
&35.19&32.56
&27.18&15.26
&42.51&40.68\\

Kalman Filter
&35.13&32.04
&26.86&15.89
&41.97&40.55\\

Latent ODE
&34.82&31.81
&27.56&15.54
&41.33&40.48\\

DSF
&\textbf{36.22}&\textbf{32.75}
&\textbf{28.51}&\textbf{16.66}
&\textbf{44.14}&\textbf{41.29}\\

\bottomrule
\end{tabular}

\caption{Downstream classification using reconstructed EEG under the $4\times$ setting.
Classifiers are trained only on ground-truth HD EEG and directly evaluated onreconstructed EEG from held-out subjects. The downstream tasks are three-class emotion recognition for SEED, four-class emotion recognition for SEED-IV, and four-class motor-imagery classification for BCI2000.}
\label{tab:downstream}
\end{table}

\begin{table}[t]
\centering
\footnotesize
% \resizebox{\columnwidth}{!}{
\begin{tabular}{l|cc|cc|cc}
\toprule
Method
& \multicolumn{2}{c|}{SEED}
& \multicolumn{2}{c|}{SEED-IV}
& \multicolumn{2}{c}{BCI2000}\\
& Acc & F1
& Acc & F1
& Acc & F1\\
\midrule

Sparse LD
&38.25&36.01
&28.96&23.68
&23.65&20.80\\

GT-HD
&42.89&41.74
&36.42&32.88
&27.80&24.60\\

\midrule

ESTFormer
&39.26&37.57
&32.77&29.58
&24.51&21.38\\

STAD
&39.74&37.23
&32.43&30.10
&24.66&21.61\\

SRGDiff
&40.82&37.85
&33.56&29.58
&25.22&22.77\\

Mamba aggregation
&40.37&37.70
&32.77&29.41
&24.56&21.64\\

Kalman Filter
&40.42&37.79
&31.96&29.18
&23.86&20.69\\

Latent ODE
&39.03&38.12
&32.26&30.07
&24.66&21.61\\

DSF
&\textbf{41.55}&\textbf{39.93}
&\textbf{34.84}&\textbf{31.63}
&\textbf{26.61}&\textbf{23.65}\\

\bottomrule
\end{tabular}
\caption{Compatibility with a frozen EEG foundation encoder under the $4\times$ setting. A linear classifier is fitted on frozen representations of ground-truth HD EEG from the training split and then fixed for all reconstruction methods. Higher accuracy and macro-F1 indicate better functional compatibility with the pretrained EEG representation.}
\label{tab:foundation_model}
\end{table}

\subsection{Experimental Setup}

\paragraph{Datasets and protocol.}

We evaluate DSF on SEED, SEED-IV, and BCI2000, containing 62, 62, and 64 EEG channels, respectively. EEG signals are manually segmented into consecutive 4-second windows whose temporal order is preserved within each recording to construct causal EEG streams. We consider three nominal reconstruction scales, $2\times$, $4\times$, and $8\times$, corresponding to 32, 16, and 8 visible electrodes, respectively. Each scale uses a predefined fixed sparse-electrode layout, which is shared by all compared methods and remains unchanged across consecutive windows under the clean-layout protocol.
We adopt a subject-independent evaluation protocol with disjoint training, validation, and test data. Validation is used for model selection, while the test set is reserved for final evaluation. We report NMSE, PCC, and reconstruction SNR on unobserved electrodes as the primary reconstruction metrics. Detailed electrode layouts and evaluation settings are provided in Appendix~\ref{app:implementation}.

\paragraph{Baselines.}

We compare DSF with both EEG-specific reconstruction methods and general temporal/state-estimation baselines. The EEG-specific group includes spherical-spline interpolation, ESTFormer \cite{li2025estformer}, STAD \cite{wang2025stad}, MASER \cite{maser2024enhancing}, and SRGDiff \cite{liu2026srgdiff}. We further compare with Mamba-based temporal aggregation \cite{dao2024mamba}, a learned latent-space Kalman filtering baseline \cite{kalman1960new}, and Latent ODE \cite{rubanova2019latent} under the same causal stream protocol. For controlled temporal comparisons, these baselines use the same window-level sparse observation and mask inputs and the same waveform-decoding protocol, while replacing the temporal state propagation mechanism with Mamba aggregation, latent Kalman filtering, or ODE-based state evolution, respectively. All learnable methods use the same train/validation/test partitions and sparse observation layouts; temporal baselines receive the same available history as DSF.

\subsection{Main Reconstruction Results}

We first evaluate the overall reconstruction performance of DSF under clean sparse observations. Table~\ref{tab:main_results} compares DSF with EEG-specific reconstruction methods and temporal/state-estimation baselines on SEED, SEED-IV, and BCI2000 across three reconstruction scales.
DSF achieves the strongest overall reconstruction performance among the evaluated methods. Its advantage is retained as the reconstruction scale increases, although absolute reconstruction quality decreases for all approaches when fewer electrodes are observed. The consistent improvement across datasets and sparsity levels suggests that history-driven prediction and innovation-based correction provide complementary information for reconstructing unobserved electrodes. DSF also maintains practical computational efficiency, achieving sub-millisecond inference latency while remaining faster than the evaluated EEG-specific neural reconstruction methods and Mamba aggregation (Table~\ref{tab:efficiency}).

\subsection{Robustness under Unreliable Observations}

We further evaluate DSF when current sparse observations are degraded by measurement noise, additional channel dropout, or their combination. These perturbations simulate practical cases in which visible electrodes become noisy or temporarily unavailable.
As shown in Table~\ref{tab:robustness_uncertainty}, DSF retains the best reconstruction performance under all three perturbation settings. Together with its clean-setting performance in Table~\ref{tab:main_results}, these results show that the advantage of DSF is preserved when the current observations become noisy or partially unavailable. The historical-context analysis in Appendix~\ref{app:temporal_stability} further shows that sequential state propagation remains stable as additional preceding windows are incorporated, without evident error accumulation over the tested sequence lengths.

\subsection{Task-oriented Evaluation of Reconstructed EEG}

Signal-level reconstruction metrics do not directly measure whether recovered EEG remains useful for downstream decoding. We therefore evaluate reconstructed signals through downstream classification and their compatibility with a frozen pretrained EEG encoder.
For downstream classification, a separate lightweight EEGNet classifier is trained for each dataset using only ground-truth high-density EEG from the training subjects. The trained classifiers are then kept fixed and directly evaluated on reconstructed EEG from held-out subjects.
SEED and SEED-IV evaluate three-class and four-class emotion recognition, respectively, while BCI2000 evaluates four-class motor-imagery classification.
The classifiers are not adapted to reconstructed signals, so the evaluation reflects how well task-relevant information is preserved after reconstruction.

We further evaluate whether reconstructed EEG remains compatible with a pretrained EEG representation model. We use EEGPT \citep{wang2024eegpt}, a pretrained Transformer encoder for EEG representation learning. The EEGPT encoder is kept frozen, and a linear classification head is fitted only on frozen representations extracted from ground-truth HD EEG in the training split. The resulting encoder and classifier are then fixed when evaluating reconstructed EEG from different methods. A smaller performance gap to GT-HD EEG therefore indicates better functional compatibility with the pretrained representation space.
As shown in Table~\ref{tab:downstream} and
Table~\ref{tab:foundation_model}, DSF achieves the strongest downstream
performance among the evaluated reconstruction methods and consistently
narrows the gap to GT-HD EEG.
These results show that the reconstruction advantage of DSF extends beyond
point-wise signal fidelity to task-relevant information, while its improved
performance with the frozen encoder further demonstrates compatibility with
pretrained EEG representations.

\subsection{Ablation of Dynamic State Filtering}

We further isolate the contributions of the two reconstruction pathways and
the key components of innovation filtering.
We first evaluate the dynamical and innovation-filtering branches independently
to examine whether their predictions provide complementary information.
Within the filtering branch, we then remove the state-conditioned routing term,
the topology bias, or the uncertainty-aware correction, respectively.
Table~\ref{tab:ablation} compares these variants with the complete DSF.
The branch-only variants quantify the contribution of combining history-driven
prediction with observation-driven filtering, while the component ablations
evaluate the roles of adaptive innovation routing, electrode topology, and
uncertainty-aware state correction.

\begin{table}[t]
\centering
\begin{tabular}{lccc}
\toprule
Variant & NMSE$\downarrow$ & PCC$\uparrow$ & SNR$\uparrow$ \\
\midrule
Dynamical branch only
& 0.094074 & 0.929406 & 10.2334 \\
Innovation-filtering branch only
& 0.087592 & 0.925794 & 10.5059 \\
w/o state-conditioned routing
& 0.077795 & 0.945703 & 11.9904 \\
w/o topology bias
& 0.085433 & 0.934917 & 11.9341 \\
w/o uncertainty correction
& 0.086127 & 0.933842 & 12.8716 \\
\midrule
Full DSF
& \textbf{0.063584} & \textbf{0.968340} & \textbf{13.8554} \\
\bottomrule
\end{tabular}

\caption{
Ablation study on BCI2000 under the $2\times$ reconstruction setting.
Branch-only variants evaluate the complementarity of history-driven prediction
and innovation filtering, while component ablations isolate the effects of
state-conditioned routing, electrode topology, and uncertainty-aware state
correction.
}
\label{tab:ablation}
\end{table}

\subsection{Visualization of Spatial Pattern Preservation}

\begin{figure*}[t]
    \centering
    \includegraphics[width=0.8\textwidth]{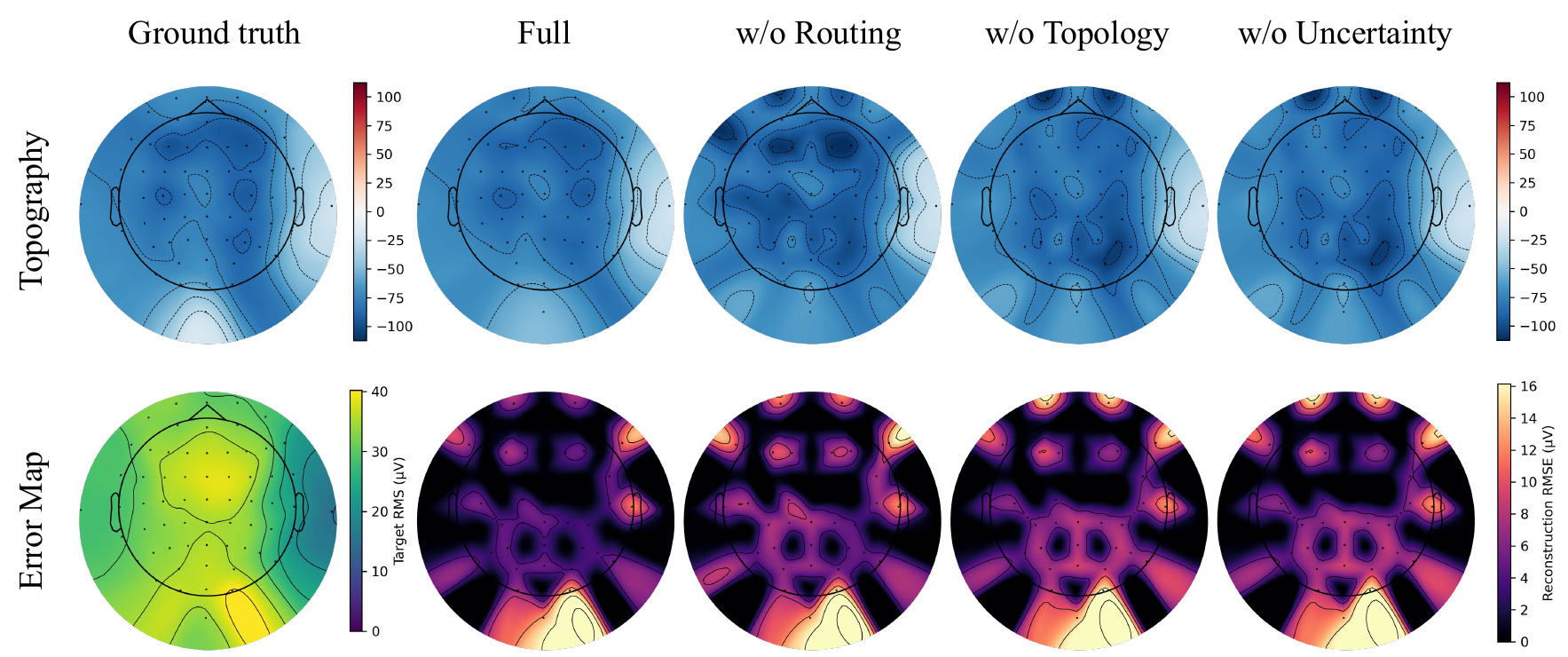}
    \caption{
    Spatial reconstruction under the $2\times$ setting.
    The first row shows the ground-truth scalp potential distribution and
    reconstructions from different methods, while the second row shows their
    corresponding channel-wise reconstruction errors.
    }
    \label{fig:topology}
\end{figure*}

To complement the signal-level metrics, we visualize scalp potential
distributions and channel-wise reconstruction errors under the $2\times$
setting.
As shown in Fig.~\ref{fig:topology}, DSF more closely reproduces the
ground-truth spatial potential pattern and yields lower reconstruction errors
across electrodes.
This qualitative result is consistent with the quantitative gains and suggests
that topology-aware innovation propagation helps preserve spatial EEG
structure at unobserved locations.

\section{Conclusion}

In this work, we rethink EEG spatial super-resolution from static signal reconstruction to dynamic neural state filtering. 
Instead of directly mapping sparse observations to dense EEG signals, we formulate reconstruction as sequential state estimation, where neural dynamics are decomposed into predictable evolution and observation-induced innovation. 
Based on this formulation, we propose Dynamic State Filtering (DSF), which propagates historical neural states to obtain predictive priors, selectively propagates observation-driven corrections through a state-conditioned transition process constrained by spatial topology, and performs uncertainty-aware state correction before waveform reconstruction.
Extensive experiments demonstrate that DSF improves reconstruction fidelity, preserves task-relevant neural representations, and maintains robustness under dynamically changing and unreliable observations.
These results suggest that explicitly modeling neural state evolution provides a promising direction for EEG spatial super-resolution and other incomplete sensing problems involving continuously evolving physiological signals.

\bibliographystyle{iclr2027_conference}
\bibliography{iclr2027_conference}

\appendix
\clearpage

\section{LLM Usage}
Large Language Models (LLMs) were used to aid in the writing and polishing of the manuscript. Specifically, we used an LLM to assist in refining the language, improving readability, and ensuring clarity in various sections of the paper. The model helped with tasks such as sentence rephrasing, grammar checking, and enhancing the overall flow of the text.

It is important to note that the LLM was not involved in the ideation, research methodology, or experimental design. All research concepts, ideas, and analyses were developed and conducted by the authors. The contributions of the LLM were solely focused on improving the linguistic quality of the paper, with no involvement in the scientific content or data analysis.

The authors take full responsibility for the content of the manuscript, including any text generated or polished by the LLM. We have ensured that the LLM-generated text adheres to ethical guidelines and does not contribute to plagiarism or scientific misconduct.

\section{Statistical Stability Analysis}

To evaluate the stability and reproducibility of DSF, we report the mean and standard deviation over repeated experiments.
All experiments are conducted with three independent training runs using different random seeds.
All runs follow the same subject-independent train/test split, where test subjects are completely excluded during training.
For each run, reconstruction performance is first averaged over all test subjects and all EEG windows from the corresponding test recordings.
The reported standard deviation is then computed across the three independent runs, reflecting variation caused by random initialization and optimization stochasticity rather than subject-level differences.
For the clean-layout experiments, the predefined sparse-electrode layout is fixed across runs and shared by all compared methods. For observation-perturbation experiments, only the additional corruption realizations, such as channel dropout and observation noise, are independently sampled across runs.
Normalization statistics are computed only from the training subjects and are shared across all runs.

\subsection{Variance of Reconstruction Performance}

To evaluate the stability of the reconstruction capability under different sparsity levels, we report the variance of DSF under clean-layout reconstruction settings.
Specifically, for each reconstruction scale, we use the same predefined sparse-electrode layout as in the main experiments, and this layout remains fixed across all repeated runs. Dense EEG recordings are used as reconstruction targets.
For each reconstruction scale, the model is independently trained three times with different random seeds and evaluated on all test subjects.
Table~\ref{tab:variance_main} reports the mean and standard deviation of NMSE, PCC, and SNR across repeated runs.

The results show that DSF exhibits consistently low variance across different reconstruction scales and datasets.
The standard deviations of all metrics remain small compared with the corresponding mean values, indicating that the reconstruction performance is not sensitive to random initialization and optimization stochasticity.

In particular, the variation remains limited even under higher reconstruction scales, where the reconstruction task becomes more challenging.
This demonstrates that the proposed dynamic state decomposition and reliability-aware update mechanism provide stable optimization behavior rather than relying on specific training trajectories.

\subsection{Variance under Observation Perturbations}

To investigate whether the robustness improvement of DSF is consistent under different observation degradations, we further report the variance under unreliable measurement conditions.
Following the robustness evaluation protocol, we consider three perturbation settings: noisy visible channels, additional channel dropout, and their combination.
For each perturbation setting, the same perturbation strength is applied to all compared methods, while random perturbation realizations are independently sampled across runs.
Table~\ref{tab:variance_robustness} reports the mean and standard deviation of NMSE and PCC for all compared methods.

The results demonstrate that DSF maintains stable performance under different observation perturbations.
Compared with other methods, DSF achieves both better reconstruction accuracy and comparable or smaller performance variations across repeated perturbation realizations.

This suggests that the reliability-aware state writing mechanism can effectively suppress the influence of unreliable observations and maintain consistent latent state evolution under stochastic measurement degradation.

\begin{table*}[t]
\centering
\small
\caption{
Mean and standard deviation of reconstruction performance under the clean-layout protocol.
Each entry is reported as mean (first line) and standard deviation (second line) over three independent runs.
}
\label{tab:variance_main}
\begin{tabular}{ll|ccc|ccc|ccc}
\toprule
\multirow{2}{*}{Method}
&
\multirow{2}{*}{Scale}
&
\multicolumn{3}{c|}{SEED}
&
\multicolumn{3}{c|}{SEED-IV}
&
\multicolumn{3}{c}{BCI2000}
\\
&
&
NMSE$\downarrow$
&
PCC$\uparrow$
&
SNR$\uparrow$
&
NMSE$\downarrow$
&
PCC$\uparrow$
&
SNR$\uparrow$
&
NMSE$\downarrow$
&
PCC$\uparrow$
&
SNR$\uparrow$
\\
\midrule

DSF
&
$2\times$
&
\makecell{0.2009\\$\pm$0.0016}
&
\makecell{0.8858\\$\pm$0.0004}
&
\makecell{7.5341\\$\pm$0.0389}
&
\makecell{0.2120\\$\pm$0.0008}
&
\makecell{0.8766\\$\pm$0.0006}
&
\makecell{7.4504\\$\pm$0.0186}
&
\makecell{0.0636\\$\pm$0.0016}
&
\makecell{0.9683\\$\pm$0.0007}
&
\makecell{13.8554\\$\pm$0.0311}
\\

DSF
&
$4\times$
&
\makecell{0.3875\\$\pm$0.0027}
&
\makecell{0.7812\\$\pm$0.0008}
&
\makecell{4.4382\\$\pm$0.0118}
&
\makecell{0.3925\\$\pm$0.0013}
&
\makecell{0.7593\\$\pm$0.0005}
&
\makecell{4.4527\\$\pm$0.0097}
&
\makecell{0.1334\\$\pm$0.0006}
&
\makecell{0.9251\\$\pm$0.0003}
&
\makecell{10.5268\\$\pm$0.0313}
\\

DSF
&
$8\times$
&
\makecell{0.6672\\$\pm$0.0120}
&
\makecell{0.5854\\$\pm$0.0091}
&
\makecell{2.1514\\$\pm$0.1053}
&
\makecell{0.5346\\$\pm$0.0003}
&
\makecell{0.6610\\$\pm$0.0003}
&
\makecell{2.9690\\$\pm$0.0053}
&
\makecell{0.2075\\$\pm$0.0009}
&
\makecell{0.8846\\$\pm$0.0004}
&
\makecell{8.1209\\$\pm$0.0252}
\\

\bottomrule
\end{tabular}
\end{table*}

\begin{table*}[t]
\centering
\small
\caption{
Mean and standard deviation of robustness evaluation under unreliable observations.
Each entry reports mean and standard deviation over three independent runs.
}
\label{tab:variance_robustness}
\begin{tabular}{l|cc|cc|cc}
\toprule
Method
&
\multicolumn{2}{c|}{Noise}
&
\multicolumn{2}{c|}{Dropout}
&
\multicolumn{2}{c}{Noise+Dropout}
\\
&
NMSE$\downarrow$
&
PCC$\uparrow$
&
NMSE$\downarrow$
&
PCC$\uparrow$
&
NMSE$\downarrow$
&
PCC$\uparrow$
\\
\midrule

ESTFormer
&
\makecell{0.5469\\$\pm$0.0105}
&
\makecell{0.7503\\$\pm$0.0022}
&
\makecell{0.5959\\$\pm$0.0098}
&
\makecell{0.6976\\$\pm$0.0012}
&
\makecell{0.6148\\$\pm$0.0070}
&
\makecell{0.6838\\$\pm$0.0022}
\\

STAD
&
\makecell{0.5945\\$\pm$0.0584}
&
\makecell{0.7411\\$\pm$0.0043}
&
\makecell{0.6385\\$\pm$0.0484}
&
\makecell{0.6850\\$\pm$0.0021}
&
\makecell{0.6690\\$\pm$0.0468}
&
\makecell{0.6653\\$\pm$0.0036}
\\

SRGDiff
&
\makecell{0.4906\\$\pm$0.0136}
&
\makecell{0.7524\\$\pm$0.0011}
&
\makecell{0.6918\\$\pm$0.0566}
&
\makecell{0.6305\\$\pm$0.0108}
&
\makecell{0.7237\\$\pm$0.0603}
&
\makecell{0.6134\\$\pm$0.0119}
\\

Mamba Aggregation
&
\makecell{0.4369\\$\pm$0.0039}
&
\makecell{0.7504\\$\pm$0.0009}
&
\makecell{0.5251\\$\pm$0.0071}
&
\makecell{0.6817\\$\pm$0.0028}
&
\makecell{0.5516\\$\pm$0.0080}
&
\makecell{0.6812\\$\pm$0.0031}
\\

Kalman Filter
&
\makecell{0.4685\\$\pm$0.0043}
&
\makecell{0.7246\\$\pm$0.0026}
&
\makecell{0.5568\\$\pm$0.0016}
&
\makecell{0.6521\\$\pm$0.0011}
&
\makecell{0.5862\\$\pm$0.0011}
&
\makecell{0.6435\\$\pm$0.0012}
\\

Latent ODE
&
\makecell{0.4493\\$\pm$0.0070}
&
\makecell{0.7238\\$\pm$0.0023}
&
\makecell{0.5379\\$\pm$0.0073}
&
\makecell{0.6759\\$\pm$0.0039}
&
\makecell{0.5579\\$\pm$0.0095}
&
\makecell{0.6693\\$\pm$0.0054}
\\

DSF
&
\makecell{\textbf{0.4014}\\$\pm$0.0052}
&
\makecell{\textbf{0.7727}\\$\pm$0.0018}
&
\makecell{\textbf{0.4933}\\$\pm$0.0045}
&
\makecell{\textbf{0.7078}\\$\pm$0.0026}
&
\makecell{\textbf{0.5048}\\$\pm$0.0071}
&
\makecell{\textbf{0.6998}\\$\pm$0.0043}
\\

\bottomrule
\end{tabular}
\end{table*}

\section{Detailed Model Architecture and Topology Configuration}
\label{app:model_details}

\begin{table*}[t]
\centering
\small
\caption{Implementation details of DSF.}
\label{tab:dsf_implementation}
\begin{tabular}{ll}
\toprule
Component & Setting \\
\midrule

Input channels
& SEED/SEED-IV: 62; BCI2000: 64 \\

Window length
& 800 samples (4 s) \\

Sampling rate
& 200 Hz \\

Stream length
& $L=6$ \\

Dynamical latent dimension
& 128 \\

Filtering feature dimension
& 32 \\

Euler steps
& $S=4$ \\

Propagation interval
& $d_t=1.0$ normalized model-time interval \\

Dynamics module
& 2-layer Conv1D velocity field + GELU \\

Dynamics kernel size
& 3 \\

Innovation routing
& State-conditioned masked query--key routing \\

Correction uncertainty
& 2-layer shared MLP + Softplus \\

State update gain
& $K_k=P_k^-/(P_k^-+R_k+\epsilon)$ \\

Uncertainty state
& Channel-wise propagated surrogate $P_k\in\mathbb{R}^{C}$ \\

Decoder
& ConvTranspose1D($k=4,s=2$)$\times2$ + Conv1D($k=5$) \\

Topology coordinates
& MNE \texttt{standard\_1005} montage \\

Topology scale
& Median non-zero pairwise electrode distance \\

STFT
& $n_{\mathrm{fft}}=128$, window=128, hop=32 \\

Loss weights
& $\lambda_{\mathrm{rec}}=1.0$,
$\lambda_{\mathrm{vis}}=1.0$,
$\lambda_{\mathrm{spec}}=0.2$,
$\lambda_{\mathrm{unc}}=10^{-3}$ \\

Batch size
& 4 \\

Learning rate
& $10^{-3}$ \\

Weight decay
& $10^{-4}$ \\

\bottomrule
\end{tabular}
\end{table*}

\subsection{Dynamical Branch Configuration}

For each EEG window, the sparse waveform
$O_k\in\mathbb{R}^{C\times T}$ and the broadcast visibility mask are
concatenated along the channel dimension, resulting in an input of size
$2C\times T$.
The observation encoder consists of two strided temporal convolutional layers
with kernel sizes $7$ and $5$, both using stride $2$.
For the $T=800$ samples used in the reconstruction experiments, the encoder
produces
\begin{equation}
Z_k\in\mathbb{R}^{128\times 200}.
\end{equation}
Accordingly, the propagated and updated dynamical states satisfy
\begin{equation}
H_k^-,H_k^+\in\mathbb{R}^{128\times 200}.
\end{equation}
The same GRUCell parameters are shared across the 200 latent temporal
positions.

The continuous dynamics operator $\Phi_\theta$ is implemented as a two-layer
Conv1D velocity field with GELU activation. One transition between consecutive
EEG windows is represented by a normalized model-time interval of length
$d_t=1.0$ and numerically integrated using four Euler substeps of size
$1/4$. These substeps are numerical integration steps and should not be
interpreted as physical one-second EEG intervals.

The dynamical decoder consists of two transposed-convolution layers followed
by a Conv1D projection, mapping the updated latent state back to
$\mathbb{R}^{C\times T}$.

\subsection{Innovation-routing Configuration}

The predictive waveform state
$S_k^-\in\mathbb{R}^{C\times T}$ and the observed innovation
$E_k\in\mathbb{R}^{C\times T}$ are processed independently by temporal
encoders whose parameters are shared across electrodes.
The resulting feature tensors satisfy
\begin{equation}
F_k^p,F_k^e\in\mathbb{R}^{C\times d\times T},
\qquad d=32.
\end{equation}
Temporal averaging yields electrode-wise representations in
$\mathbb{R}^{C\times d}$ for query--key computation, while the innovation
value representation retains the temporal dimension.

\subsection{Electrode Topology Construction}

Electrode positions are obtained from the standard 10--05 scalp montage
(\texttt{standard\_1005}) according to the channel names and ordering of each
dataset.
Before pairwise distances are computed, the 3-D template coordinates are
standardized along each coordinate axis.
We define
\begin{equation}
d_{ij}
=
\left\|
\tilde p_i-\tilde p_j
\right\|_2
\end{equation}
and construct the topology bias as
\begin{equation}
B_{ij}
=
-\frac{d_{ij}^{2}}{2\sigma_p^{2}},
\end{equation}
where $\sigma_p$ is set to the median non-zero pairwise electrode distance.
Hence $B_{ij}\leq 0$, and distant electrode pairs receive a stronger spatial
penalty.
For SEED and SEED-IV, CB1 and CB2 are mapped to the posterior template
locations P9 and P10, respectively, because the corresponding labels are not
available in the standard montage.

\subsection{State Initialization and Fusion Configuration}

At the beginning of each input sequence, the recurrent states of both
branches are re-initialized. For the dynamical branch, the latent state is
initialized by the encoded first sparse observation. For the
innovation-filtering branch, the initial posterior waveform state is set
directly to the first sparse observation,
\begin{equation}
S_1^+ = O_1.
\end{equation}
The corresponding uncertainty state is initialized from a learnable
channel-wise parameter $p_0\in\mathbb{R}^{C}$ as
\begin{equation}
P_1^+
=
\operatorname{softplus}(p_0)+\epsilon,
\end{equation}
where $p_0$ is initialized to zero and $\epsilon=10^{-6}$.
The state is then propagated causally within the input sequence.

For bounded fusion, we set $\alpha_{\max}=0.20$. The channel-wise fusion
weights are parameterized as
\begin{equation}
\alpha_c=\alpha_{\max}\sigma(a_c).
\end{equation}
All fusion logits are initialized such that the filtering branch initially
contributes $0.05$, giving
\begin{equation}
a_c
=
\operatorname{logit}\!\left(
\frac{0.05}{0.20}
\right)
=
\log\frac{1}{3}
\approx -1.0986.
\end{equation}
During the final fusion stage, the parameters of both reconstruction branches
are frozen and only the channel-wise fusion logits $\{a_c\}_{c=1}^{C}$ are
optimized.

\section{Additional Experimental Details}
\label{app:implementation}

\subsection{Training Objective and Optimization}

The total training objective combines reconstruction loss, visible-channel consistency loss, spectral preservation loss, and uncertainty regularization:
\begin{equation}
\mathcal{L} = \lambda_{\mathrm{rec}}\mathcal{L}_{\mathrm{rec}} + \lambda_{\mathrm{vis}}\mathcal{L}_{\mathrm{vis}} + \lambda_{\mathrm{spec}}\mathcal{L}_{\mathrm{spec}} + \lambda_{\mathrm{unc}}\mathcal{L}_{\mathrm{unc}} .
\end{equation}

Here, $\mathcal{L}_{\mathrm{rec}}$ is the mean absolute error between the
reconstructed and ground-truth dense EEG, while
$\mathcal{L}_{\mathrm{vis}}$ computes the same absolute reconstruction error
only over currently visible electrodes.
The spectral term is defined as
\begin{equation}
\mathcal{L}_{\mathrm{spec}}
=
\left\|
|\operatorname{STFT}(\hat X)|
-
|\operatorname{STFT}(X)|
\right\|_1,
\end{equation}
where we use a Hann window with $n_{\mathrm{fft}}=128$, window length $128$,
and hop length $32$.
The first EEG segment is used for state initialization and is excluded from
these losses.

For the innovation-filtering branch, the uncertainty regularizer is
\begin{equation}
\mathcal L_{\mathrm{unc}}
=
\frac{1}{L-1}
\sum_{k=2}^{L}
\left[
\operatorname{Mean}_{c}\log(1+R_{k,c})
+
0.1\,\operatorname{Mean}_{c}
K_{k,c}(1-K_{k,c})
\right].
\end{equation}
The first term prevents unconstrained growth of the learned correction
uncertainty, while the second discourages persistently ambiguous intermediate
gains. This regularizer is applied only when training the innovation-filtering
branch; it is zero for the dynamical-branch and final fusion stages.

Unless otherwise stated, we set $\lambda_{\mathrm{rec}}=1.0$, $\lambda_{\mathrm{vis}}=1.0$, $\lambda_{\mathrm{spec}}=0.2$, and $\lambda_{\mathrm{unc}}=10^{-3}$.

Optimization is performed using AdamW with batch size 4, initial learning rate $10^{-3}$, and weight decay $10^{-4}$.
Gradients are clipped with a maximum norm of 1.0. 
Normalization statistics are computed only from the training split and applied to validation and test splits. 
Model selection is based on validation performance, while the test set is used only for final reporting.

\subsection{Implementation of Temporal State-Estimation Baselines}
\label{app:temporal_baselines}

To ensure a controlled comparison with alternative temporal modeling
strategies, we implement Mamba aggregation, Kalman filtering, and Latent
ODE under the same causal EEG-stream protocol used by DSF.
All three baselines receive the same zero-filled sparse observations
$O_k$ and visibility masks $M_k$, use the same training, validation,
and test partitions, and are provided with exactly the same preceding
EEG windows.

For all learnable temporal baselines, the sparse waveform and the
broadcast visibility mask are concatenated along the channel dimension.
Each 4-second observation is encoded by two temporal convolutional
layers with kernel sizes $7$ and $5$ and stride $2$, producing a
window-level latent representation
$Z_k\in\mathbb{R}^{D\times 200}$.
The temporal baselines differ only in how these representations are
propagated across consecutive EEG windows.
Their resulting latent states are decoded by the same waveform-decoder
family consisting of two transposed-convolution layers followed by a
Conv1D projection.

\paragraph{Mamba aggregation.}
We use Mamba as a causal temporal aggregator across EEG windows rather
than across raw samples within an individual window.
Specifically, for each latent temporal location $\tau$, the sequence
$\{Z_k(:,\tau)\}_{k=1}^{L}$ is processed along the window dimension by
four residual Mamba blocks with hidden dimension $D=132$.
Each block uses state dimension $16$, convolution width $4$, and
expansion factor $2$.
Consequently, the representation of window $k$ depends only on the
current and preceding EEG windows and does not access future segments.

\paragraph{Kalman Filter.}
For the Kalman baseline, we perform filtering in the encoded latent
space with state dimension $D=150$.
The first encoded observation initializes the latent state.
The predictive mean is propagated through a learned linear transition
matrix $A$,
\begin{equation}
\mu_k^- = A\mu_{k-1}^+.
\end{equation}
We use diagonal process and measurement covariance terms for efficient
filtering.
Under the identity latent observation model, the update is
\begin{equation}
K_k =
\frac{P_k^-}{P_k^-+R},
\qquad
\mu_k^+
=
\mu_k^-
+
K_k\odot(Z_k-\mu_k^-),
\end{equation}
with the corresponding diagonal covariance update
$P_k^+=(1-K_k)\odot P_k^-$.
The transition matrix and the positive process and measurement
uncertainty parameters are learned jointly with the encoder and decoder.
This baseline therefore implements a learned latent-space Kalman
filter while retaining the classical prediction--innovation--update
structure.

\paragraph{Latent ODE.}
We implement a causal ODE--RNN realization of Latent ODE for sequential
EEG reconstruction.
The latent state has dimension $D=132$.
Between consecutive EEG windows, the previous state is propagated over
a normalized model-time interval of length $1$ using a two-layer latent
velocity field and four Euler integration steps.
The propagated state is subsequently updated by the current encoded
observation using a GRUCell:
\begin{equation}
H_k^- =
\operatorname{ODESolve}(H_{k-1}^+),
\qquad
H_k^+
=
\operatorname{GRU}(Z_k,H_k^-).
\end{equation}
The ODE velocity field uses an intermediate dimension of $140$.

For all three baselines, the first EEG window is used to initialize the
temporal state and is excluded from the reconstruction losses.
The remaining windows are optimized using the same waveform
reconstruction, visible-channel consistency, and STFT-based spectral
preservation objectives used for DSF, with
$\lambda_{\mathrm{rec}}=1$,
$\lambda_{\mathrm{vis}}=1$, and
$\lambda_{\mathrm{spec}}=0.2$.
The uncertainty regularizer is specific to the DSF
innovation-filtering branch and is therefore not applied to these
baselines.
All models are trained using AdamW with the same batch size, learning
rate, weight decay, gradient clipping, normalization, and model
selection protocol.

\begin{table}[t]
\centering
\small
\caption{Implementation details of temporal/state-estimation baselines.
The temporal axis denotes the sequence of consecutive 4-second EEG
windows.}
\label{tab:temporal_baseline_details}
\begin{tabular}{lccc}
\toprule
Method & State dim. & Temporal transition & Params (M) \\
\midrule
Mamba Aggregation
& 132
& 4-layer causal Mamba
& 0.819 \\
Kalman Filter
& 150
& Learned linear filtering
& 0.364 \\
Latent ODE
& 132
& ODE propagation + GRU
& 0.463 \\
\bottomrule
\end{tabular}
\end{table}

\subsection{Evaluation Protocol and Metrics}

We adopt a subject-independent evaluation protocol with disjoint training and
test subjects.
Within the training split, 20\% of complete EEG streams are held out for
validation and used for checkpoint selection and early stopping.
Normalization statistics are computed exclusively from the remaining training
data, and the test set is accessed only after model selection for final
evaluation.

Sparse observations are generated by applying channel-visibility masks to the
full-density EEG recordings.
For each reconstruction scale, all compared methods are evaluated using the
same visible-electrode layouts.
For observation-corruption experiments, Gaussian noise with standard deviation
$0.1$ is added to the visible electrodes after signal normalization, while
channel dropout randomly removes an additional $20\%$ of the currently visible
electrodes at each window. The combined setting applies both perturbations.
Identical perturbation realizations are shared across methods to ensure paired
evaluation conditions.

We use NMSE, PCC, and reconstruction SNR as the primary signal-level metrics, computed exclusively over electrodes that are absent from the sparse input. Predictions at visible electrodes are never replaced by ground-truth signals before metric computation.

For downstream evaluation, classifiers are trained exclusively on
ground-truth HD EEG from training subjects and directly evaluated on
reconstructed EEG from held-out subjects, without adaptation to reconstruction
artifacts.
SEED and SEED-IV are evaluated using three-class and four-class emotion
recognition, respectively.

For BCI2000, we formulate the downstream task as four-class motor-imagery
classification. EEG segments are extracted from 0.5--4.0~s after cue onset
and resampled to 128~Hz, resulting in 64-channel segments with 449 temporal
samples. The 0.5--4.0~s cropping and 128-Hz resampling are applied only when
reconstructed and ground-truth signals are converted to the input format of
the downstream BCI2000 classifier. We use a subject-independent split with
68 training subjects and 20 test subjects. Ground-truth HD EEG and zero-filled
sparse LD EEG are included as upper and lower reference conditions,
respectively.

\subsection{EEG Foundation Encoder Evaluation}

For the pretrained-representation evaluation, we use EEGPT
(\textit{Pretrained Transformer for Universal and Reliable Representation of EEG Signals})
with the publicly released \texttt{braindecode/eegpt-pretrained} checkpoint
(revision \texttt{e41cb3ae2ce4fd9eb736862292c91f8128d15618}).
The encoder has a 512-dimensional embedding space with 8 Transformer layers,
8 attention heads, and 4 summary tokens, containing approximately 25.3M
encoder parameters.

The EEGPT encoder is used in evaluation mode without parameter updates.
For each downstream dataset, we first extract frozen EEGPT embeddings from
ground-truth HD EEG in the training split and fit a dataset-specific logistic-regression
classifier on these embeddings.
The fitted classifier is then kept fixed and evaluated using ground-truth HD EEG,
sparse LD EEG, and reconstructed EEG from different methods.
No reconstructed test samples or test labels are used when fitting the classifier.

The public EEGPT pretraining corpus includes SEED and PhysioNetMI.
Since the BCI2000 evaluation used in this work is derived from the PhysioNet
motor-imagery recordings, the SEED and BCI2000 evaluations may overlap with
the encoder's pretraining corpus.
Therefore, these results are used to measure compatibility with the pretrained
EEGPT representation space rather than transfer to entirely unseen datasets.
SEED-IV is not explicitly listed in the public EEGPT pretraining corpus.

\section{Analysis of Temporal Dynamics and Filtering Stability}

The core contribution of DSF lies in reformulating static EEG reconstruction as dynamic state filtering. To verify the robustness of this temporal modeling mechanism, we conduct analyses from two perspectives: the stability of state accumulation over long sequences and the sensitivity to the amount of available historical context.

\subsection{Stability of Sequential State Propagation}
\label{app:temporal_stability}

In dynamic state filtering, the amount of historical information available for reconstructing a target EEG segment depends on the length of the observed sequence. This analysis evaluates whether sequential state accumulation improves reconstruction quality without introducing error drift. Although DSF is trained using streams of length $L=6$, its recurrent state-transition operators are shared across windows and do not depend on a fixed sequence length. For this analysis, the trained model is therefore rolled out autoregressively over longer test streams without retraining or parameter modification.

We evaluate the reconstruction NMSE under different historical context lengths on the SEED dataset, where each trial contains consecutive 4-second windows.
The performance dynamics are compared against a sequential state estimation baseline, Kalman Filter, which also maintains temporal states but does not explicitly decompose state transitions into predictable evolution and observation-driven innovation.

\begin{figure}[h]
    \centering
    \includegraphics[width=\columnwidth]{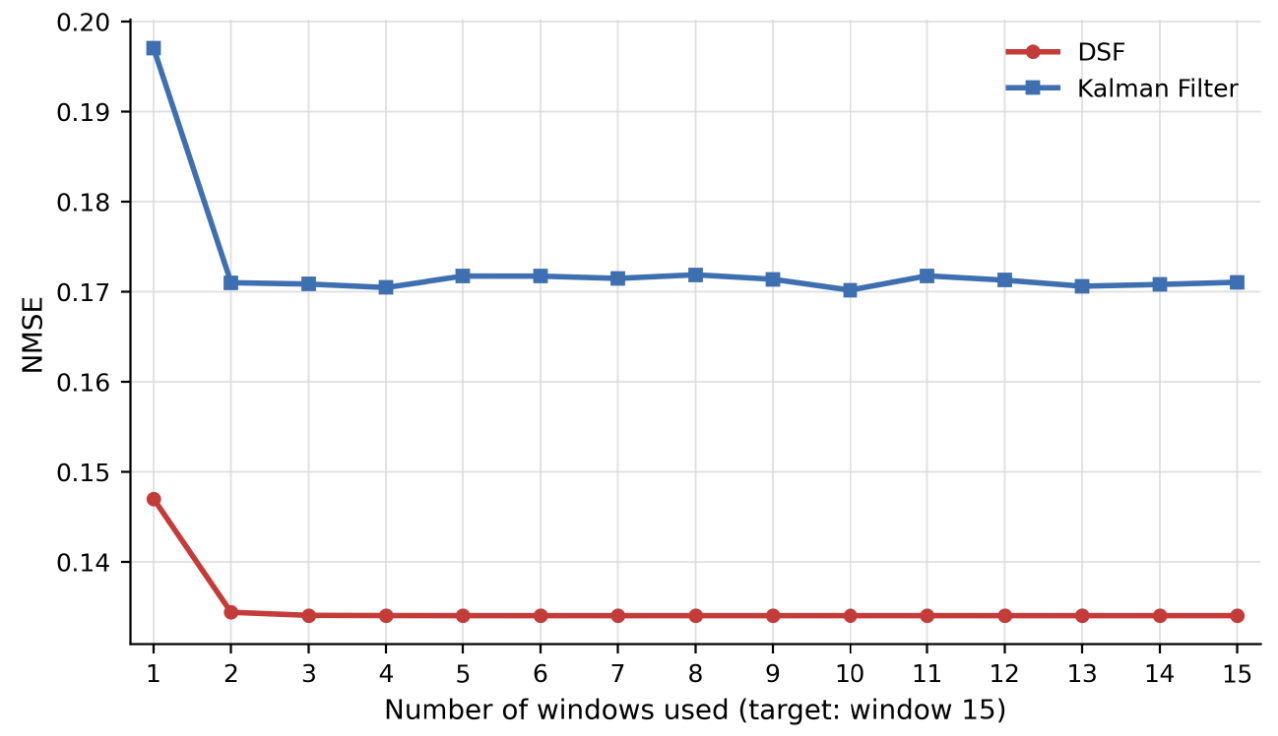}
    \caption{Reconstruction NMSE of a fixed target window under different historical context lengths. DSF improves rapidly with additional temporal context and remains stable as the available history increases.}
    \label{fig:temporal_stability}
\end{figure}

As shown in Fig.~\ref{fig:temporal_stability}, we observe two key properties of DSF that confirm the robustness of the proposed filtering mechanism:

\begin{itemize}
    \item \textbf{Rapid Context Utilization (Cold-start Robustness):} When only the target window is available ($h=1$), DSF performs reconstruction without preceding temporal observations. As the historical context increases, the reconstruction error decreases rapidly and reaches a stable regime within a short sequence length. This demonstrates that the observation-guided innovation routing can effectively extract useful temporal priors from limited historical context.
    
    \item \textbf{Long-term Stability (No Error Accumulation):} For longer historical contexts, the reconstruction performance remains stable rather than degrading. Although longer sequences introduce more accumulated states, DSF does not exhibit error drift. This behavior is consistent with stable reliability-aware state updating, which regulates the contribution of observation-driven corrections during sequential propagation.
\end{itemize}

In conclusion, DSF maintains stable reconstruction performance as historical context increases, demonstrating both rapid adaptation to limited observations and robustness against long sequential state accumulation.

\subsection{Sensitivity to Historical Context Length}

\begin{table*}[htbp]
\centering
\caption{Quantitative results with different historical context lengths.}
\begin{tabular}{c c c c}
\toprule
History Duration & NMSE $\downarrow$ & PCC $\uparrow$ & SNR $\uparrow$ \\
\midrule
2 s  & $0.1344 \pm 0.0003$ & $0.9221 \pm 0.0002$ & $10.5089 \pm 0.0043$ \\
3 s  & $0.1348 \pm 0.0005$ & $0.9220 \pm 0.0002$ & $10.5253 \pm 0.0255$ \\
4 s  & $0.1340 \pm 0.0001$ & $0.9219 \pm 0.0001$ & $10.5459 \pm 0.0074$ \\
8 s  & $0.1348 \pm 0.0003$ & $0.9219 \pm 0.0001$ & $10.4931 \pm 0.0077$ \\
16 s & $0.1352 \pm 0.0015$ & $0.9213 \pm 0.0005$ & $10.4186 \pm 0.0115$ \\
\bottomrule
\end{tabular}
\label{tab:history_length}
\end{table*}

The 4-second target window used in the main experiments is kept fixed throughout this analysis. 
To examine sensitivity to the amount of available historical context, we vary only the preceding EEG history while leaving the current target unchanged. 
For the 2-s and 3-s settings, the model input remains a 4-second (800-sample) history window: only the most recent 2 or 3 seconds immediately preceding the target are retained, while the earlier samples are set to zero. 
The 4-s setting uses the complete preceding window, while the 8-s and 16-s settings use two and four complete preceding windows, respectively.
Thus, the input dimensionality and model architecture remain unchanged across all settings.
All other model configurations and evaluation settings are kept fixed.
Experiments are conducted on the SEED dataset.
As shown in Table~\ref{tab:history_length}, DSF maintains stable reconstruction performance across different amounts of available historical context.

\begin{itemize}

\item \textbf{Effective utilization of short history:}
Even with only 2--3 seconds of historical context, DSF achieves competitive reconstruction accuracy.
This demonstrates that the proposed filtering mechanism can extract informative temporal priors from limited preceding observations, providing robustness under short-context scenarios.

\item \textbf{Stable behavior with extended history:}
Increasing the historical duration from 4 seconds to 16 seconds does not introduce noticeable performance degradation.
Although longer historical contexts introduce more sequential information into state propagation, DSF does not exhibit noticeable error drift or unstable state evolution.
This behavior is consistent with stable reliability-aware state updating, which regulates the influence of observation-driven corrections during sequential propagation.

\end{itemize}

Overall, these results demonstrate that DSF is robust to different historical context lengths.
The model can rapidly benefit from available temporal information while maintaining stable state evolution without relying on excessively long memory.

\section{State-decoding Decoupling Analysis}

A design principle of the innovation-filtering branch is to separate waveform-state correction from the final reconstruction head. To examine whether the benefit of filtering depends strongly on a particular decoder architecture, we fix the state propagation, innovation routing, and uncertainty-aware correction modules and replace only the reconstruction head.

As shown in Table~\ref{tab:decoder_compatibility}, the filtered waveform state can be decoded effectively by multiple reconstruction heads, although performance varies across decoder architectures. These results support decoder compatibility rather than decoder independence.

\begin{table}[t]
\centering
\begin{tabular}{lccc}
\toprule
Decoder & NMSE$\downarrow$ & PCC$\uparrow$ & SNR$\uparrow$ \\
\midrule
CNN Decoder
& \textbf{0.072988} & \textbf{0.962581} & \textbf{12.9605} \\
Transformer Decoder
& 0.073857 & 0.962035 & 12.8193 \\
U-Net Decoder
& 0.074725 & 0.961617 & 12.7093 \\
MLP Decoder
& 0.079244 & 0.953925 & 11.6280 \\
\bottomrule
\end{tabular}
\caption{Decoder compatibility analysis. The filtering modules of DSF are fixed while different reconstruction heads are used for waveform decoding.}
\label{tab:decoder_compatibility}
\end{table}

\subsection{Computational Efficiency}

Table~\ref{tab:efficiency} compares model size, computational cost, and
inference latency for one 4-second EEG window.
Despite maintaining sequential states and an additional innovation-filtering
pathway, DSF is faster than the evaluated EEG-specific neural reconstruction methods and Mamba aggregation, while retaining sub-millisecond inference latency.

\begin{table*}[t]
\centering
\caption{Efficiency comparison for one 4-s EEG inference step (batch size 1, FP32).}
\label{tab:efficiency}
\begin{tabular}{lrrr}
\toprule
Method & Params (M) & GFLOPs / 4 s & Runtime (ms) \\
\midrule
ESTFormer
& 0.677
& 0.490
& 1.123 $\pm$ 0.009 \\

STAD
& 0.707
& 0.916
& 1.620 $\pm$ 0.134 \\

SRGDiff
& 0.933
& 0.974
& 1.548 $\pm$ 0.048 \\

Mamba Aggregation
& 0.819
& 0.624
& 0.956 $\pm$ 0.009 \\

Kalman Filter
& 0.364
& 0.264
& 0.416 $\pm$ 0.006 \\

Latent ODE
& 0.463
& 0.299
& 0.495 $\pm$ 0.006 \\

\textbf{Ours / DSF}
& \textbf{0.562}
& \textbf{0.460}
& \textbf{0.753 $\pm$ 0.036} \\
\bottomrule
\end{tabular}
\end{table*}

\end{document}